\documentclass[conference]{IEEEtran}
\IEEEoverridecommandlockouts
\usepackage{cite}
\usepackage{amsmath,amssymb,amsfonts}
\usepackage{algorithmic}
\usepackage{graphicx}
\usepackage{textcomp}
\usepackage{subfigure}
\usepackage{hyperref} 
\usepackage[table]{xcolor}
\usepackage{multirow}
\definecolor{myblue}{HTML}{0000ee}
\usepackage{booktabs} 

\def\bng{\bngx}

\font\bngx=bang10

\def\*#1*#2{o\null{#2}{#1}}

\def\sh#1{\setbox0=\hbox{#1}%
     \kern-.02em\copy0\kern-\wd0
     \kern.04em\copy0\kern-\wd0
     \kern-.02em\raise.0433em\box0 }
\usepackage{caption} 
\def\BibTeX{{\rm B\kern-.05em{\sc i\kern-.025em b}\kern-.08em
    T\kern-.1667em\lower.7ex\hbox{E}\kern-.125emX}}

\begin{document}

\title{Motamot: A Dataset for Revealing the Supremacy of Large Language Models over Transformer Models in Bengali Political Sentiment Analysis\\

\thanks{\textsuperscript{*}These authors contributed equally to this work.}
}



\author{
\IEEEauthorblockN{Fatema Tuj Johora Faria\textsuperscript{*}, Mukaﬀi Bin Moin\textsuperscript{*}, Rabeya Islam Mumu, Md Mahabubul Alam Abir, \\
Abrar Nawar Alfy, 
Mohammad Shafiul Alam }
 Department of Computer Science and Engineering\\
 Ahsanullah University of Science and Technology (AUST), Dhaka, Bangladesh\\
 
Email-  \{fatema.faria142, mukaﬀi28, rabeya.islammomo, mahbubabir09, abraralfy49\}@gmail.com
 \\ 
\{shafiul.cse\}@aust.edu
}

\maketitle

\begin{abstract}
Sentiment analysis is the process of identifying and categorizing people’s emotions or opinions regarding various topics. Analyzing political sentiment is critical for understanding the complexities of public opinion processes, especially during election seasons. It gives significant information on voter preferences, attitudes, and current trends. In this study, we investigate political sentiment analysis during Bangladeshi elections, specifically examining how effectively Pre-trained Language Models (PLMs) and Large Language Models (LLMs) capture complex sentiment characteristics. Our study centers on the creation of the ``Motamot" dataset, comprising 7,058 instances annotated with positive and negative sentiments, sourced from diverse online newspaper portals, forming a comprehensive resource for political sentiment analysis. We meticulously evaluate the performance of various PLMs including BanglaBERT, Bangla BERT Base, XLM-RoBERTa, mBERT, and sahajBERT, alongside LLMs such as Gemini 1.5 Pro and GPT 3.5 Turbo. Moreover, we explore zero-shot and few-shot learning strategies to enhance our understanding of political sentiment analysis methodologies. Our findings underscore BanglaBERT's commendable accuracy of 88.10\% among PLMs. However, the exploration into LLMs reveals even more promising results. Through the adept application of Few-Shot learning techniques, Gemini 1.5 Pro achieves an impressive accuracy of 96.33\%, surpassing the remarkable performance of GPT 3.5 Turbo, which stands at 94\%. This underscores Gemini 1.5 Pro's status as the superior performer in this comparison.
\end{abstract}


\section{Introduction}
Political sentiment analysis \cite{pol} examines public opinion, feelings, and attitudes about political entities, events, or ideologies, which is especially important during Bangladeshi elections. Online newspapers such as Prothom Alo\footnote{\href{https://www.prothomalo.com/}{\textcolor{myblue}{https://www.prothomalo.com/}}}, Bangladesh Pratidin\footnote{\href{https://www.bd-pratidin.com/}{\textcolor{myblue}{https://www.bd-pratidin.com/}}}, Samakal\footnote{\href{https://samakal.com/}{\textcolor{myblue}{https://samakal.com/}}}, and others are important forums for political conversation and information dissemination. These platforms provide individuals with election-related news, opinions, and analyses, which influence public attitude in Bangladesh's changing political scenario.

The evolution of pre-trained language models has revolutionized NLP, offering exceptional performance across diverse tasks with minimal fine-tuning. Tailored for Bengali, models like BanglaBERT \cite{BanglaBert}, SahajBERT, and mBERT \cite{mbert} have advanced Bengali NLP applications significantly. However, their dependence on large annotated datasets poses challenges due to Bengali's limited representation in NLP. The rise of LLMs showcases remarkable accuracy without extensive training, yet models like LLaMA and GPT-3/4 face scrutiny for opaque parameters and potential misinformation. To address this, the Reinforcement Learning from Human Feedback (RLHF) mechanism \cite{Intro3} aims to ensure truthful responses, yet in Bengali, LLM application remains largely unexplored due to data scarcity.

Significant progress has been achieved in sentiment analysis across a variety of areas, including Reddit market sentiment \cite{RedditXiang}, student feedback \cite{studentNasim}, product reviews \cite{productManal}, restaurant reviews \cite{RestaurantEhsanur}, and general sentiment around the COVID-19 vaccination \cite{covid}. These attempts have made major improvements to understanding public opinion and assessing feelings in a variety of circumstances. Despite this improvement, there is still a significant gap: a scarcity of thorough research devoted only to political sentiment analysis. 

This study aims to investigate public sentiments about politics on Bangladeshi online newspapers during elections. We analyze a large amount of content from these websites to understand people's perspectives on political issues such as parties, policy, and elections. In addition, we investigate the emotion conveyed by political parties. We intend to give insight for both political parties and voters by understanding the views and opinions expressed by individuals, as well as the sentiments of political parties, allowing them to make informed decisions about which party to support. To the best of our knowledge, this is the first analysis of LLMs in the domain of ``Political Sentiment Analysis" in Bengali language. Here is a summary of the outcomes of our experiments: 
\begin{itemize}
    \item Developed a novel dataset named ``Motamot," containing 7,058 data points labeled with Positive and Negative sentiments, tailored specifically for Political Sentiment Analysis in the Bengali language. The dataset comprises 4,132 instances labeled as Positive and 2,926 instances labeled as Negative sentiments.
    
    \item Conducted comprehensive evaluations of both PLMs (BanglaBERT, Bangla BERT Base, XLM-RoBERTa, mBERT, and SahajBERT) along with LLMs (Gemini 1.5 Pro and GPT 3.5 Turbo).
    
    \item Identified that zero-shot performance of LLMs generally lags behind State-of-the-Art (SOTA) fine-tuned PLMs across most evaluation tasks, revealing substantial performance disparities among LLMs. This underscores the conclusion that current LLMs are not well-suited for addressing low-resourced language tasks in Bengali, particularly in zero-shot scenarios.

    \item Illustrated that Few-shot learning outperforms PLMs, highlighting its potential as a more effective approach for Bengali Political Sentiment Analysis tasks. Additionally, while hallucination occurred in zero-shot scenarios, Few-shot learning did not exhibit such hallucination.
\end{itemize}

\section{Related Works}
In their study, Xiang et al. \cite{RedditXiang} propose a semi-supervised approach for market sentiment analysis, utilizing LLMs to generate weak labels for Reddit posts. They incorporate Chain-of-Thought (COT) reasoning to enhance label stability and accuracy. Despite being trained on weakly labeled data, the experimental results demonstrate competitive performance against supervised models. Additionally, Zarmeen et al. \cite{studentNasim} present a hybrid sentiment analysis model for student feedback, integrating TF-IDF, N-gram, and lexicon-based features. Their study showcases superior performance over other methods and APIs, highlighting its relevance in educational contexts. By leveraging machine learning and domain-specific lexicons, the approach enables accurate sentiment analysis, aiding educators in enhancing teaching methodologies and decision-making. Meanwhile, Manal et al. \cite{productManal} explore e-commerce sentiment analysis, emphasizing the vast internet data necessary for understanding customer sentiments. Through comparisons of supervised ML models using TF-IDF, N-gram, and lexicon-based approaches, prevalent positivity is revealed. Logistic regression excels post extensive text prep and evaluation, showcasing its efficacy in predicting customer recommendations. Their findings underscore sentiment analysis' critical role in e-commerce decisions, emphasizing the need to tackle challenges like spotting fake reviews. In another study, Ehsanur et al. \cite{RestaurantEhsanur} analyze Bangladeshi food delivery app reviews using NLP, comparing AFINN, RoBERTa, and DistilBERT models. Despite challenges like limited data and noise, DistilBERT achieves the highest accuracy (77\%), highlighting its effectiveness. The study underscores the significance of sentiment analysis in the food delivery sector, suggesting the need for context-specific models to address natural language complexities. Lastly, Muntasir et al. \cite{Muntasir} develop hybrid CNN-LSTM models with various Word Embeddings to detect emotions from Bangla texts. Achieving 90.49\% accuracy and 92.83\% F1 score with Word2Vec embedding, their study aims to accurately identify happiness, anger, and sadness emotions, contributing to Bangla language sentiment analysis.

\section{Background Study}

\subsection{Pre-trained Language Models on Bangla}
Developing language-specific models presents significant hurdles, especially for languages like Bengali that have limited resources. Despite this, recent improvements have resulted in an increase in pretrained language models gaining popularity. These models have demonstrated SOTA performance across diverse downstream tasks. In this context, the efficacy of such models is briefly explored below.
\subsubsection{\textbf{ELECTRA base}}

ELECTRA by Google employs ``replaced token detection" to distinguish real tokens from substitutes, enhancing both comprehension and computational efficiency. By focusing on altered tokens, it refines context understanding and language semantics during training. BanglaBERT \cite{BanglaBert} generator, a derivative of ELECTRA, utilizes masked language modeling (MLM) on extensive Bengali corpora for pre-training.

\subsubsection{\textbf{BERT base}}
BERT (Bidirectional Encoder Representations from Transformers) is a cutting-edge pre-trained language model created by Google researchers. It transformed NLP by proposing a bidirectional method to context comprehension. Unlike prior models that processed text sequentially, BERT evaluates both the left and right contexts at the same time, capturing deeper semantic significance. Bangla BERT Base \cite{banglabertbase} and mBERT \cite{mbert} are pre-trained Bengali language models based on BERT's pioneering mask language modeling framework. 

\subsubsection{\textbf{ALBERT large}}

ALBERT emphasizes flexibility and performance, shrinking model size and computational requirements without sacrificing efficiency. It surpasses conventional BERT models through strategies like parameter reduction and layer parameter sharing. ALBERT's Lite design, featuring parameter sharing and factorized embedding, maintains quality with fewer parameters. SahajBERT\footnote{\href{https://huggingface.co/neuropark/sahajBERT}{\textcolor{myblue}{https://huggingface.co/neuropark/sahajBERT}}}, an ALBERT variant, is trained for Bengali using MLM.
\subsubsection{\textbf{RoBERTa base}}
XLM-RoBERTa \cite{roberta}, an extension of RoBERTa, is a multilingual model trained on an extensive corpus covering over 100 languages, enabling it to process diverse information sources. Operating through unsupervised learning, it autonomously learns from vast amounts of text data without human labeling. Employing masked language modeling, it predicts missing elements within text, fostering a deep understanding of word and concept relationships. Additionally, XLM-RoBERTa possesses automatic language detection capabilities, facilitating seamless multilingual processing without external cues.

\subsection{Large Language Models on Bangla}
\subsubsection{\textbf{GPT-3.5 Turbo}}
GPT-3.5 Turbo \cite{GPT} has considerable advances in comprehending and executing instructions, making it ideal for activities that need particular formatting or outputs, such as creative content development. Its fine-tuning capability enables developers to adjust its behavior to specific requirements, hence improving performance for a variety of apps. For example, the model can be fine-tuned to consistently employ a specific language or to simplify cues for desirable replies. These advancements elevate GPT-3.5 Turbo to the top of its series, providing a cost-effective and adaptable solution for text generating workloads. With a broad context window of 16,385 tokens and enhanced precision in formatting, it successfully tackles encoding concerns for non-English language functions while providing speedy answers, limited at 4,096 output tokens.
\subsubsection{\textbf{Gemini 1.5 Pro}}
Gemini 1.5 Pro \cite{Gemini}, a member of Google DeepMind's Gemini series, is a  cutting-edge multi-modal model adept at processing text, audio, and video, expanding its utility across various tasks. Its standout feature lies in its ability to grasp long-context information, fostering nuanced understanding and insightful responses. While specifics on multilingual capabilities are somewhat limited, Gemini Pro likely excels in processing text across multiple languages. Boasting a substantial input token limit of 30,720, an output token limit of 2,048, along with stringent safety measures and a 60 requests per minute rate limit, it emerges as a versatile tool for diverse linguistic endeavors, ensuring both efficiency and effectiveness.
\section{Corpus Creation}
\subsection{\textbf{Data Collection}}
We called the dataset ``Motamot" \cite{dataset} in Bengali ({\bng ``mtamt''}) and in English (Opinion).  It was meticulously compiled from a range of online newspapers focusing on political events and conversations during Bangladeshi elections. Our data collection process involved scraping articles and opinion pieces from reputable news sources, ensuring a diverse and representative sample of political discourse. ``Motamot" gives a broad look into the many opinions and conversations that shape Bangladesh's political environment.

\begin{figure*}
    \centering
    \subfigure[Train Dataset]
    {
        \includegraphics[width=2.0in]{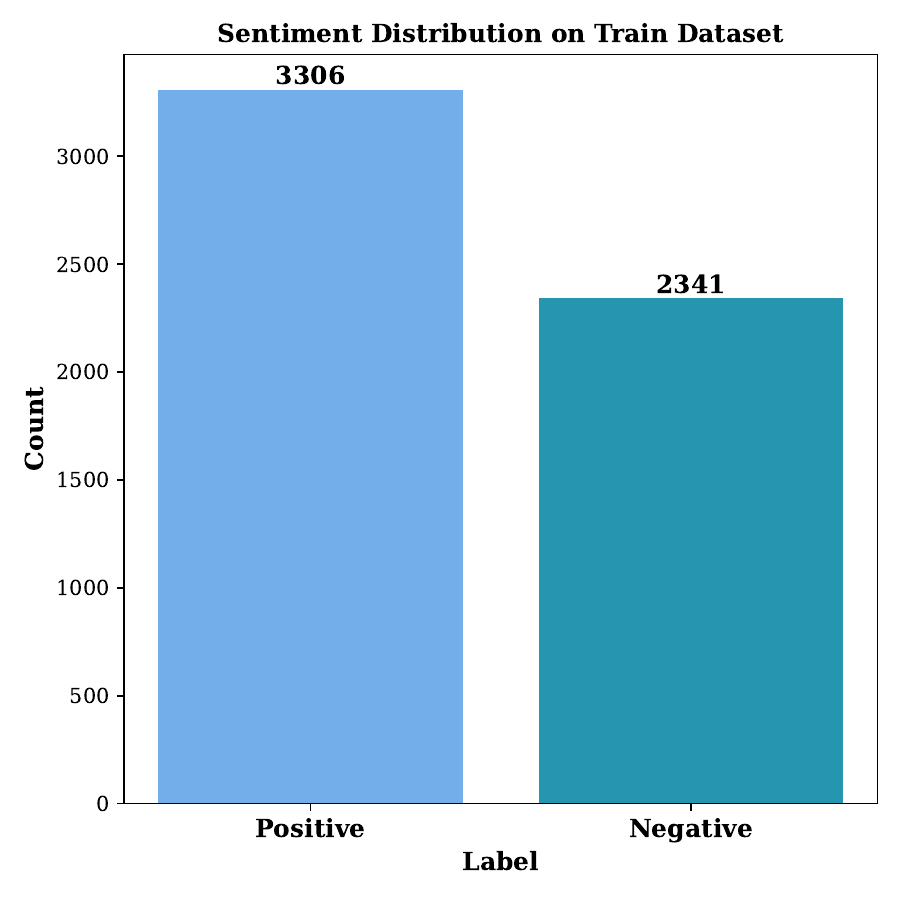}
        \label{fig:first_sub}
    }
    \subfigure[Test Dataset]
    {
        \includegraphics[width=2.0in]{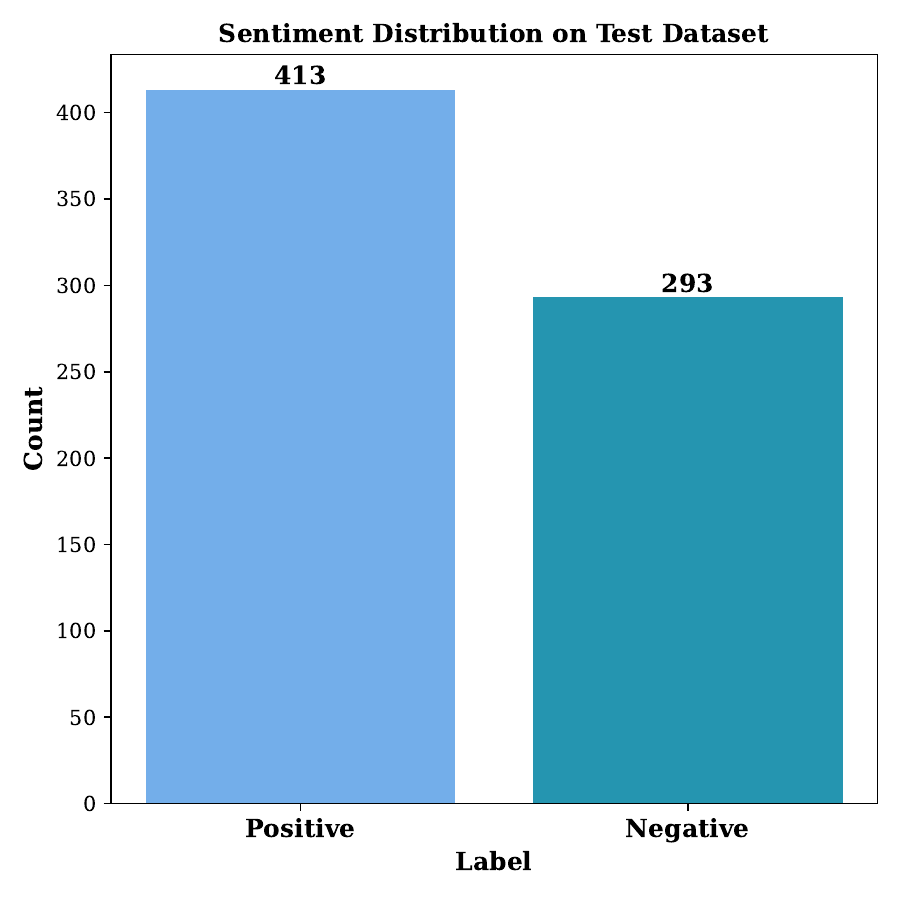}
        \label{fig:second_sub}
    }
    \subfigure[Validation Dataset]
    {
        \includegraphics[width=2.0in]{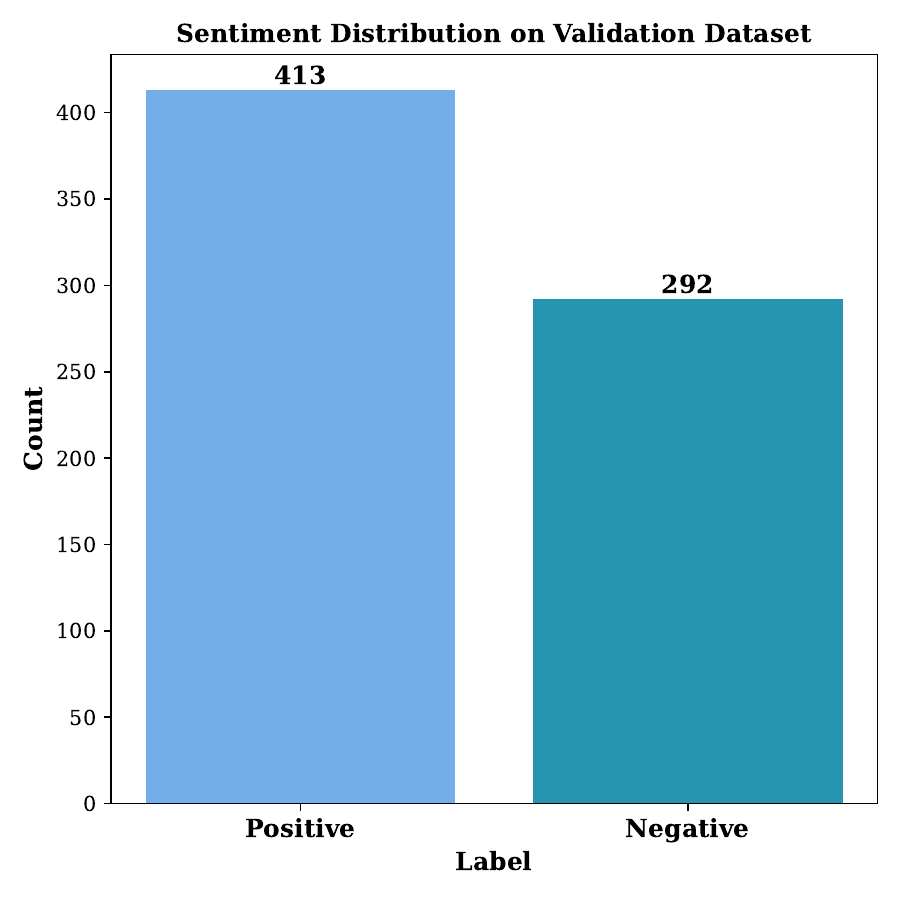}
        \label{fig:third_sub}
    }
    \caption{This figure provides a visual representation of the Analysis on the Distribution of Political Sentiments Across Train, Test, and Validation Datasets}
    \label{Sats}
\end{figure*}
\subsection{\textbf{Data Attributes}}
The dataset comprises several key attributes essential for comprehensive analysis. These attributes include the ``source\_link," providing the URL of the article or news source, alongside the ``newspaper\_name," denoting the origin of the article. The ``published\_date" attribute indicates the date of article publication, offering temporal context. Each article is accompanied by a ``headline", serving as a concise title or summary of its content. Additionally, the ``short\_description" attribute provides a brief excerpt summarizing the main points or arguments presented in the article. Finally, the ``sentiment" attribute assigns a sentiment label (e.g., Positive, Negative) to each article, facilitating sentiment analysis and classification.
\subsection{\textbf{Annotation Process}}
The dataset was extensively annotated manually, with a dedicated team of six student annotators overseeing the process. Annotators carefully analyzed each article's content to provide proper emotion labels. This entailed a careful examination of the tone, language complexities, and contextual subtleties within the articles, assuring proper classification of the opinions stated regarding political issues.
\subsection{\textbf{Annotation Guideline}}
The development of annotation guidelines was an intensive process designed to ensure uniformity and precision in sentiment sentiment labeling. These guidelines were meticulously crafted to provide annotators with clear criteria for identifying sentiment expressions within the text. Examples were included to illustrate each sentiment category, guiding annotators in their understanding and application of sentiment labels. For instance, instances expressing support, agreement, or optimism towards political figures or policies were categorized as ``Positive," while those conveying criticism, disagreement, or dissatisfaction were labeled as ``Negative." This guidance helped maintain uniformity across annotations and ensured that sentiments were accurately captured and categorized within the dataset.
\subsection{\textbf{Dataset Statistics}}
Figure \ref{Sats} shows a complete overview of the dataset's composition and scope, highlighting precise statistics across multiple subsets that have been thoroughly partitioned for optimal analysis. The dataset is divided into three essential categories: train (80\%), test (10\%), and validation (10\%).

        
        
        

\subsection{\textbf{Annotation Quality Control Process}}
The Fleiss Kappa \cite{cohenkappa} inter-rater reliability coefficient was employed with six annotators to evaluate the annotation approach's consistency. Achieving a high kappa score of 0.87 would indicate strong agreement among annotators regarding sentiment labeling, reflecting the method's reliability and uniformity. This level of agreement suggests that the quality control techniques implemented were effective in ensuring uniformity and dependability in sentiment labeling across the dataset.
\subsection{\textbf{Challenges Faced and Limitations}}
The data collection process faced challenges including biased news reporting, sentiment ambiguity, and language variations. Careful navigation was required to maintain dataset integrity, with meticulous data selection criteria and precise annotation guidelines implemented. Quality control measures were employed to mitigate these challenges' impact. The dataset includes only ``Positive" and ``Negative" sentiment labels, potentially limiting sentiment analysis granularity. Gender, location, and political biases were noted and considered during annotation, though addressing them comprehensively remains challenging. 
\subsection{\textbf{Availability and Usage}}
The ``Motamot" \cite{dataset} dataset is available in CSV format, making it easily accessible and compatible with a wide range of research tools and platforms.

\section{Implementation Details}
\subsection{Political Sentiment Analysis using PLMs}
We provide a thorough method for detecting subtle sentiment patterns in political conversation, which includes preparing the data, model fine-tuning, and performance evaluation. This approach provides a solid foundation for analyzing public opinion processes in the field of politics. Our research implementations are accessible to the public at: \href{https://github.com/Mukaffi28/Bengali-Political-Sentiment-Analysis}{\textcolor{myblue}{GitHub}}\footnote{\href{https://github.com/Mukaffi28/Bengali-Political-Sentiment-Analysis}{\textcolor{myblue}{https://github.com/Mukaffi28/Bengali-Political-Sentiment-Analysis}}}
\textbf{\\Step 1) Text Preprocessing:}
Our dataset  ``Motamot" undergoes preprocessing to ensure compatibility with PLMs. A series of normalization steps are applied, specifically designed for Bengali text. These steps include handling whitespace, commas, URLs, Unicode characters, and correcting quotation marks throughout the dataset. \\ 
\textbf{Step 2) Fine-tuning Procedure:}
During the fine-tuning process, the initialized PLMs are trained on the Political Sentiment Analysis dataset using transfer learning techniques. Model parameters are updated via gradient descent optimization algorithms, with the objective of minimizing a suitable loss function, such as cross-entropy loss. We employ the AdamW optimizer and CrossEntropyLoss as the loss function. \\
\textbf{Step 3) Training Settings:}
Hyperparameters play a crucial role in shaping the learning process and influencing model performance. Key hyperparameters such as learning rate, batch size, and epochs are adjusted during fine-tuning to optimize performance and prevent overfitting. Table \ref{tableHyper} provides detailed guidance for hyperparameter optimization to enhance model performance. \\
\textbf{Step 4) Performance Assessment:}
The performance of fine-tuned PLMs is evaluated on a held-out test set using predefined evaluation criteria. Metrics such as Accuracy, Precision, Recall, and F1 score are utilized to objectively assess model performance. Table \ref{tableBert} presents comprehensive performance metrics, on the effectiveness of fine-tuned models for Political Sentiment Analysis tasks.
\begin{figure*}
\centering
\includegraphics[width=470pt, height=300pt]{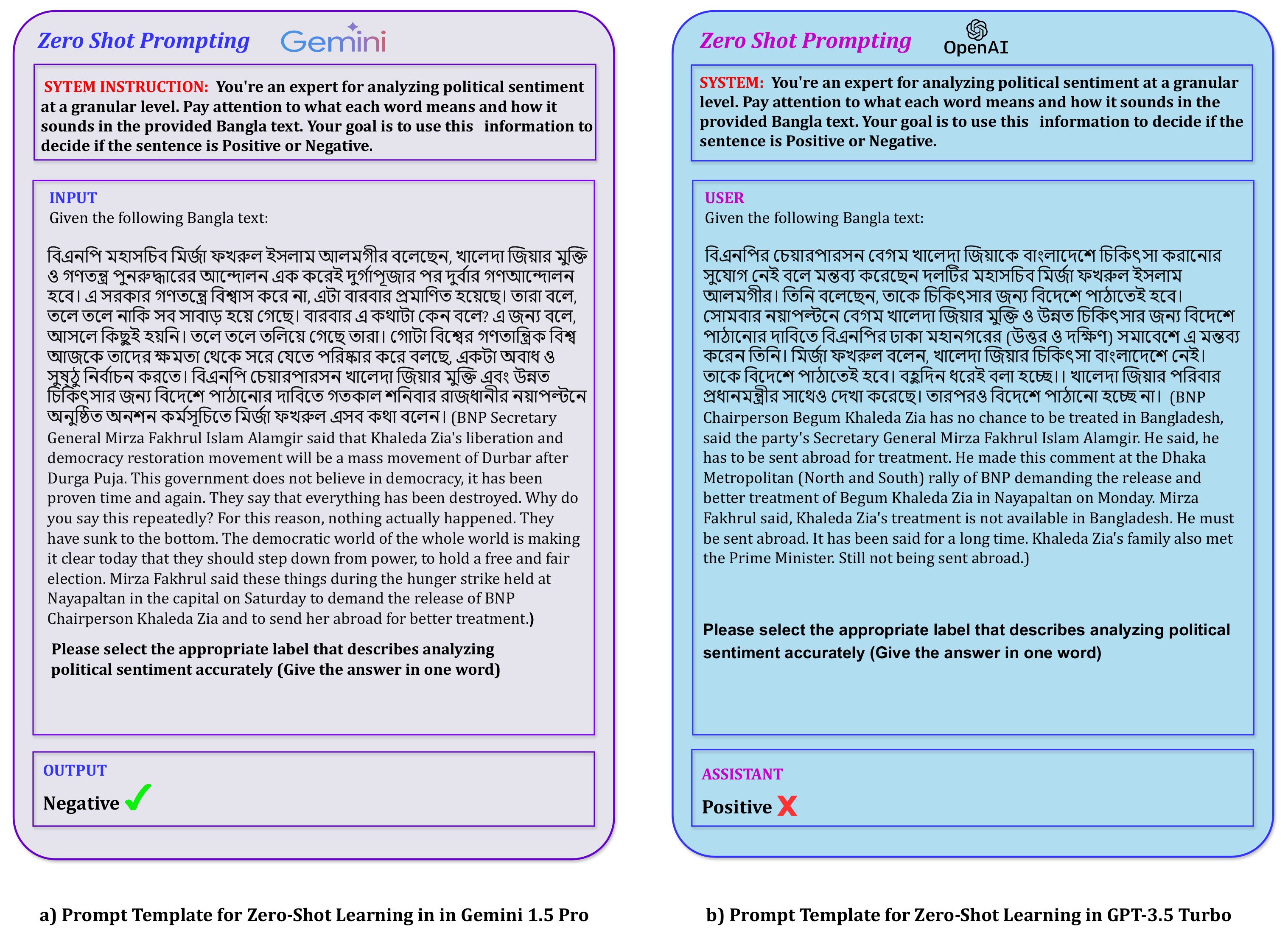}
\caption{An illustration demonstrating the key components of the prompt template designed for zero-shot learning in large language models. The template includes designations for Gemini 1.5 Pro, highlighting System Instruction, Input, and Output, as well as ChatGPT 3.5 Turbo, which highlights System, User, and Assistant interactions}\label{fig:workflow}
\end{figure*} 
\subsection{Political Sentiment Analysis using LLMs}
\subsubsection{\textbf{Data Selection}}
We randomly selected 300 data points from our ``Motamot" dataset to assess the effectiveness of Zero Shot and Few Shot prompts. Specifically, we focused on data instances with a ``short\_description" column containing more than 25 words. Each label category, including Positive and Negative sentiments, comprises 100 instances. These samples were extracted from the training subset of the ``Motamot" dataset. Considering the cost implications associated with using the OpenAI GPT-3.5 Turbo API and Gemini 1.5 Pro API, we opted to limit our experiments to this subset of the dataset rather than the entire corpus.
\subsubsection{\textbf{Prompting Template}}
In LLMs, a prompt directs text generation by specifying input parameters, crucial for task customization and output refinement. Its form varies based on the intended task, guiding LLM behavior effectively.
\begin{itemize}
    \item \textbf{Zero-shot prompting for Political Sentiment Analysis:}
    
    Zero-shot prompting in Political Sentiment Analysis involves guiding the model to predict the sentiment label (Positive or Negative) associated with a given text, without prior training on such sentiments. For instance, in a Positive sentiment scenario, the text may express support or agreement with a political figure or policy. Conversely, in a Negative sentiment scenario, the text may convey criticism or dissatisfaction. The model aims to accurately classify the sentiment based on the textual content alone. Figure \ref{fig:workflow} displays Zero-shot prompting with ``Motamot" dataset, showing premise-text pairs labeled Positive or Negative. Additionally, it highlights Gemini 1.5 Pro's Zero-shot accuracy in Political Sentiment Analysis without explicit training.

    \item \textbf{Few-shot prompting for Political Sentiment Analysis:}
    
   In Political Sentiment Analysis, few-shot prompting is a technique aimed at enhancing a model's ability to discern sentiment in text with minimal training data. In a 5-shot scenario, the model is provided with just five labeled examples, allowing it to grasp the nuances of sentiment expression. Similarly, in a 10-shot scenario, the model's exposure to labeled data doubles, refining its understanding further. Extending on this concept in a 15-shot scenario, the model is provided with fifteen labeled samples, providing a wider range of emotion expressions for learning. With each increase in shots, the model's ability to generalize and properly predict emotion in new, unlabeled texts improves. This strategy finds a compromise between data efficiency and prediction performance, allowing the model to make more accurate judgments about political emotions as it fulfills more labeled samples.

\end{itemize}

\section{Result Analysis \& Discussion}

\begin{figure*}
    \centering
    \subfigure[BanglaBERT]
    {
        \includegraphics[width=2.1in]{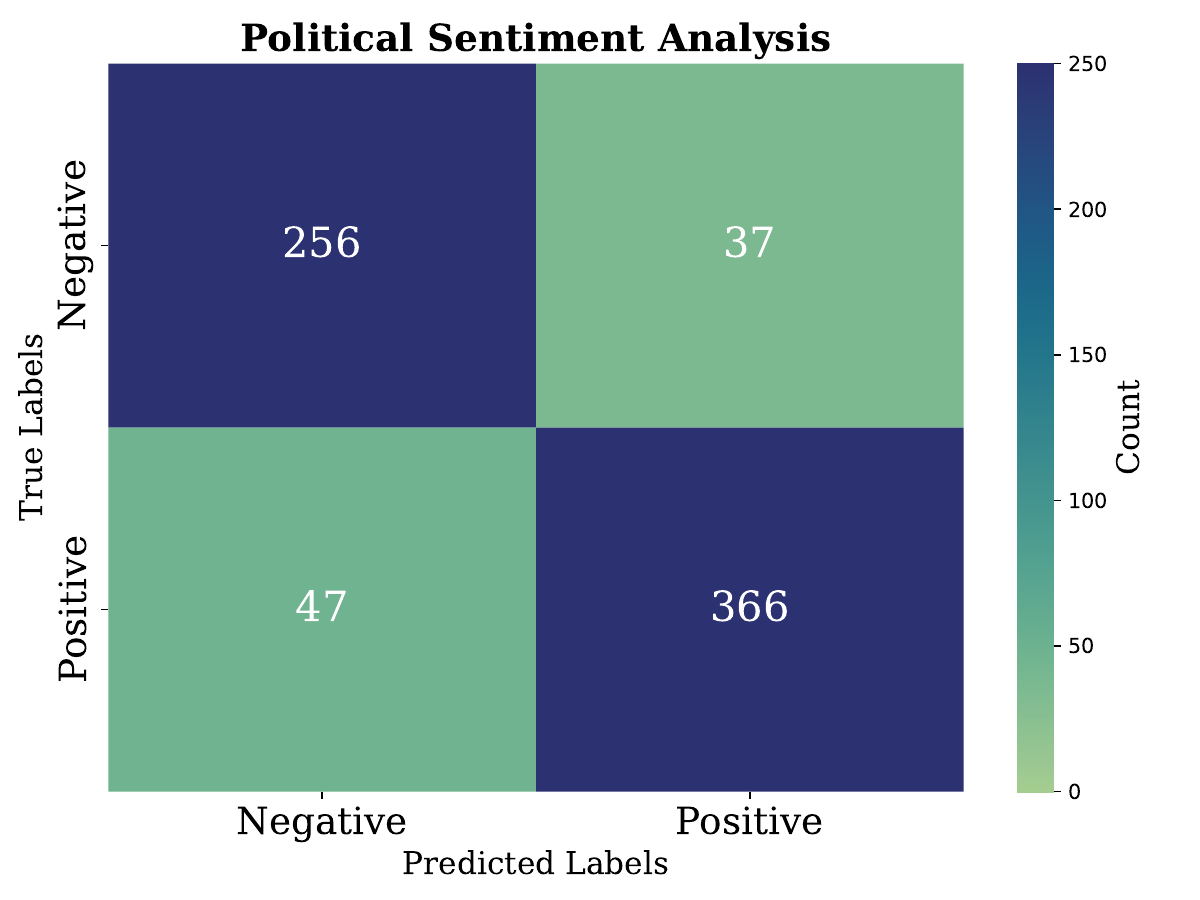}
        \label{fig:first_sub}
    }
    \subfigure[Bangla BERT Base]
    {
        \includegraphics[width=2.1in]{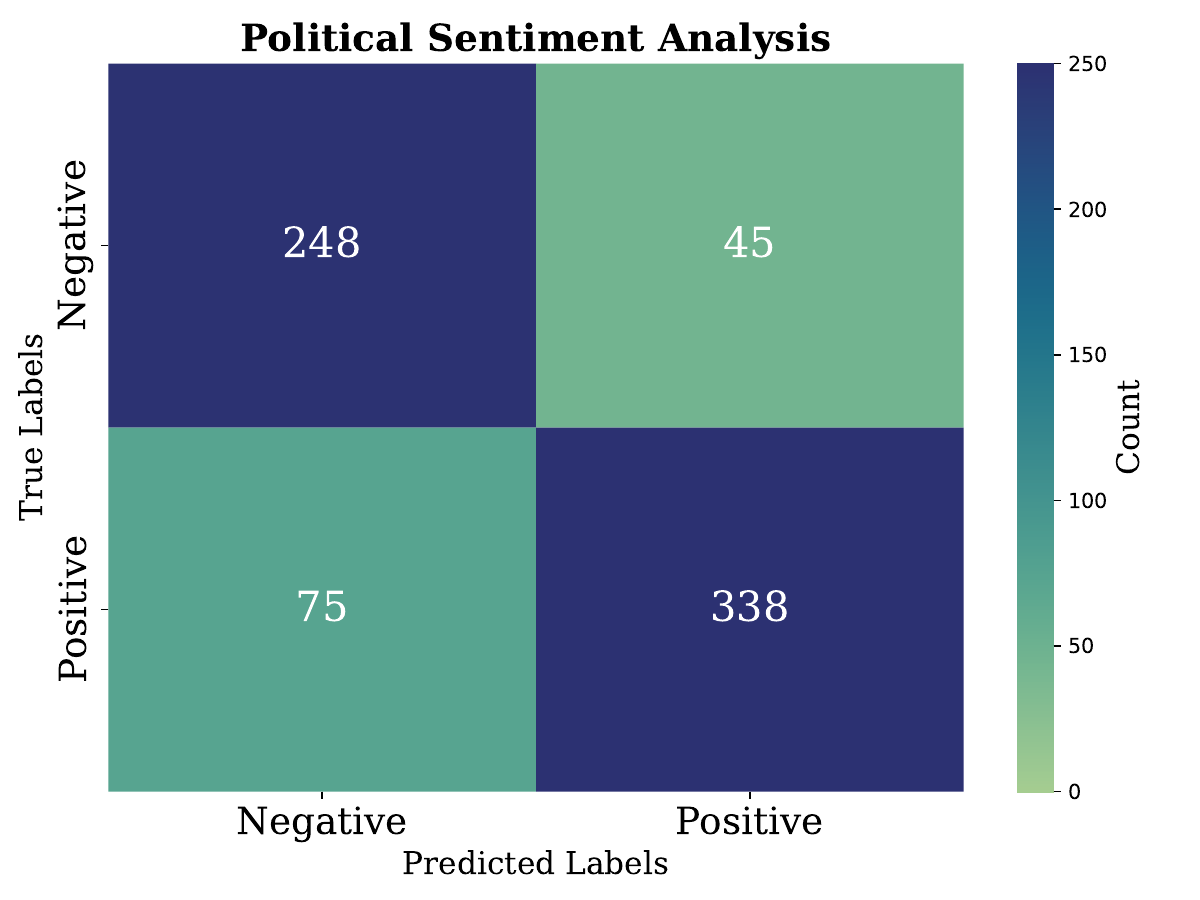}
        \label{fig:second_sub}
    }
    \subfigure[XLM-RoBERTa]
    {
        \includegraphics[width=2.1in]{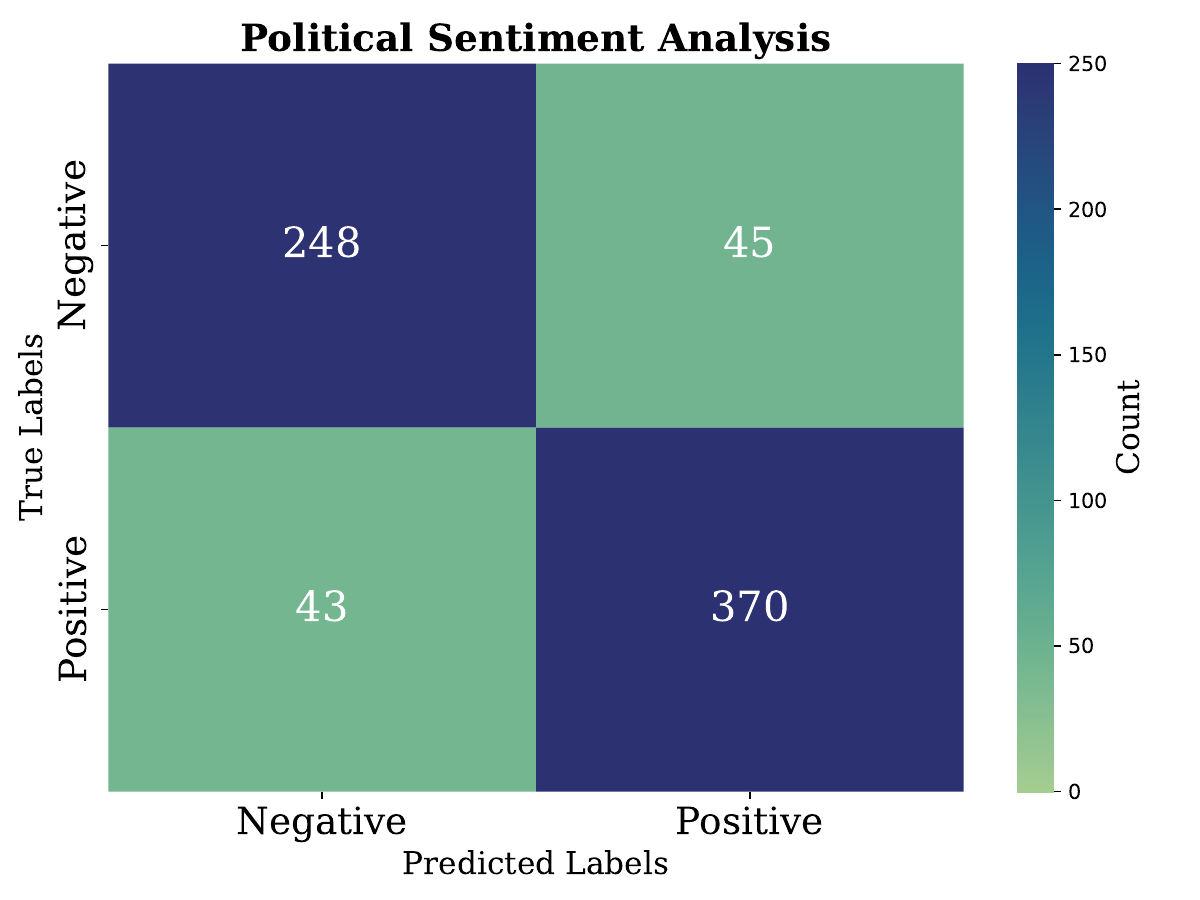}
        \label{fig:third_sub}
    }
        \subfigure[mBERT]
    {
        \includegraphics[width=2.1in]{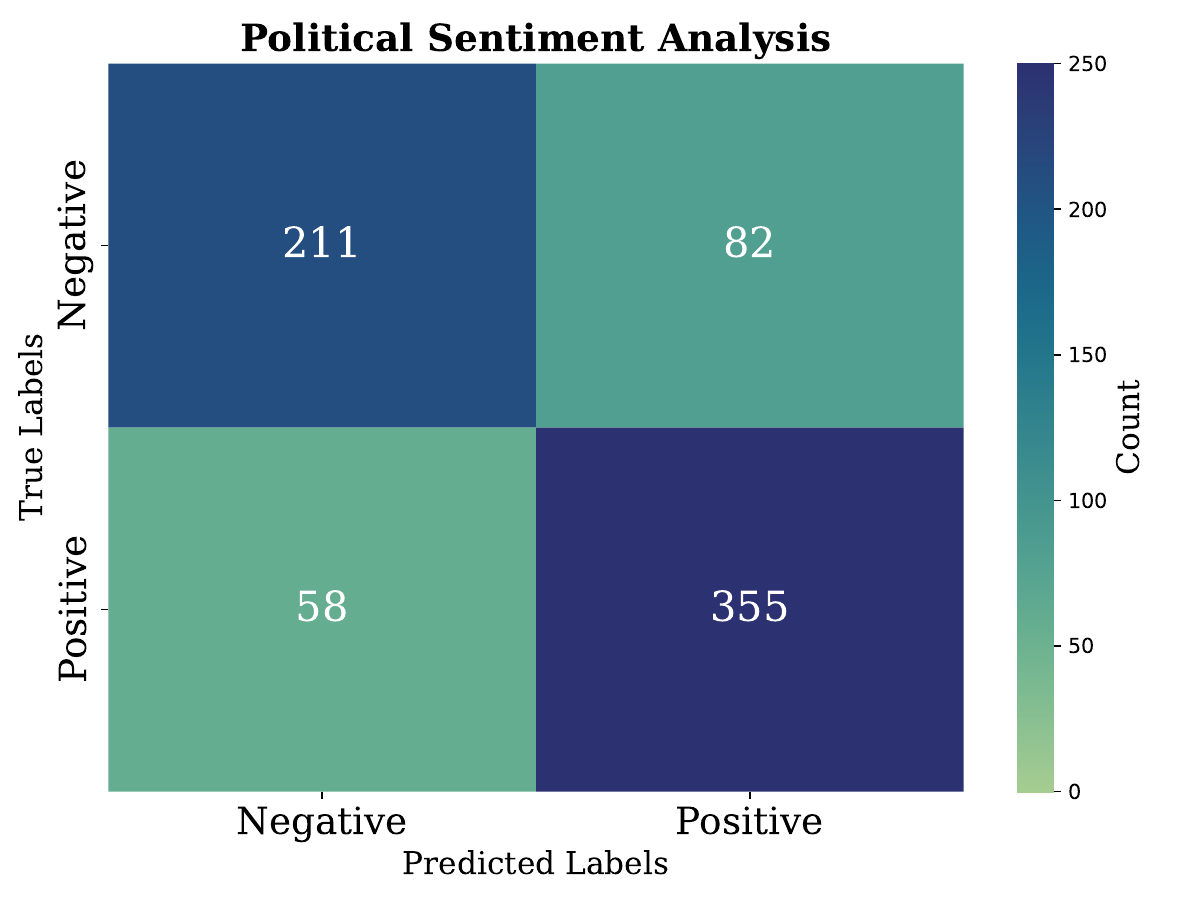}
        \label{fig:second_sub}
    }
    \subfigure[sahajBERT]
    {
        \includegraphics[width=2.1in]{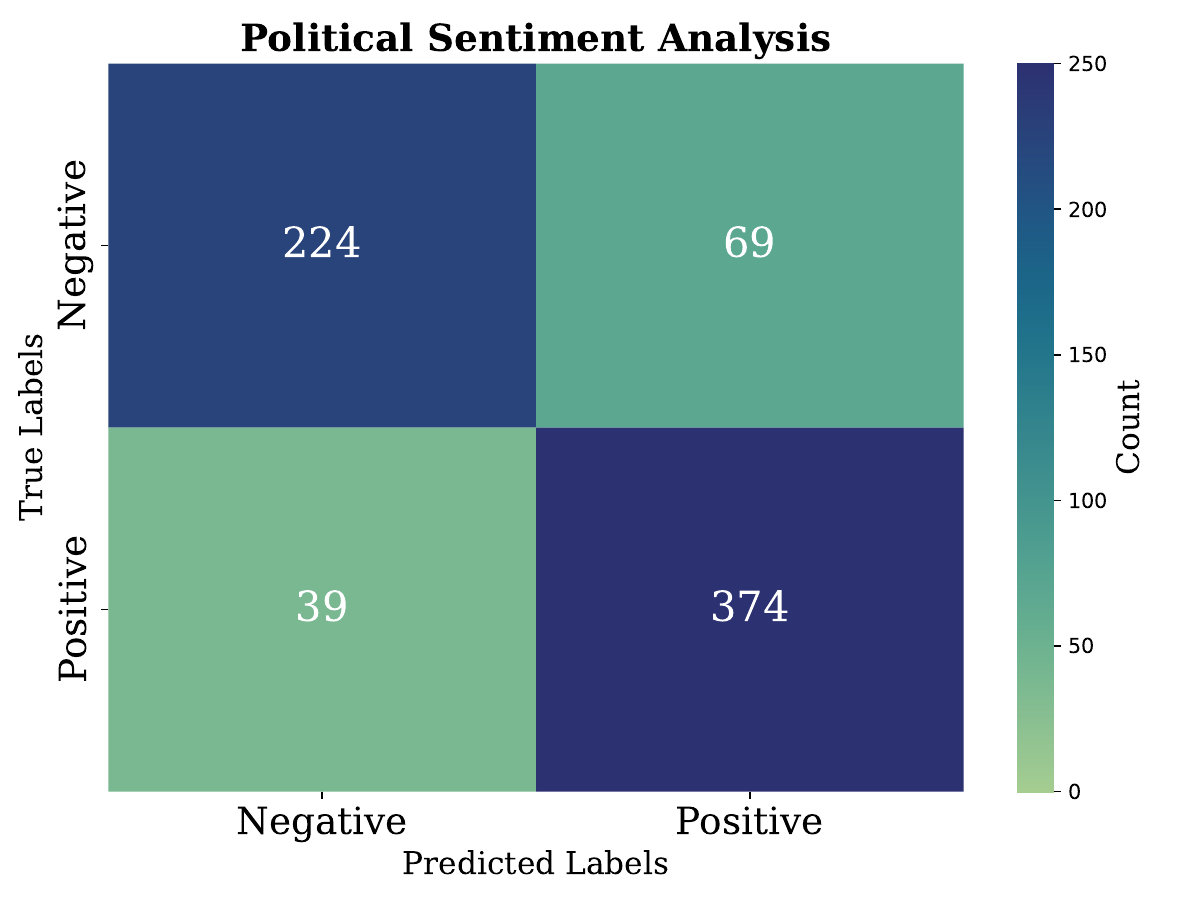}
        \label{fig:third_sub}
    }
    \caption{Visualization of confusion matrices  showing the performance of BanglaBERT, Bangla BERT Base, XLM-RoBERTa, mBERT, and sahajBERT pre-trained PLMs in political sentiment analysis. Each subfigure displays the models' classification accuracy across sentiment categories, revealing useful information about their strengths and limitations in sentiment prediction}
    \label{confusion}
\end{figure*}

\begin{table}[h]
\caption{Hyperparameter Optimization for Diverse Pre-trained Language Models}
\label{tableHyper}
\centering
\begin{tabular}{cccc}
\toprule
\textbf{Model} & \textbf{Learning Rate} & \textbf{Batch Size} & \textbf{Epochs} \\
\midrule
BanglaBERT & 2e-5 & 8 & 20 \\
Bangla BERT Base & 2e-4 & 8 & 15 \\
XLM-RoBERTa & 2e-5 & 8 & 20 \\
mBERT & 1e-5 & 8 & 20 \\
sahajBERT & 2e-5 & 6 & 15 \\
\bottomrule
\end{tabular}
\end{table}

\begin{table}[h]
\caption{Performance Metrics Comparison of Various Pre-trained Language Models Across Different Evaluation Criteria}
\label{tableBert}

  \centering  
    \begin{tabular}{ccccc}
 \toprule
        \textbf{Model} & \textbf{Accuracy} & \textbf{Precision} & \textbf{Recall} & \textbf{F1-Score} \\
        \midrule
      \textbf{BanglaBERT} & \textbf{0.8810} & \textbf{0.8765} & \textbf{0.8799} & \textbf{0.8780} \\
        
       Bangla BERT Base & 0.8300 & 0.8251 & 0.8324 & 0.8272 \\
       
        XLM-RoBERTa & 0.8753 & 0.8718 & 0.8711 & 0.8715\\
        
         mBERT &  0.8016 & 0.7983 & 0.7898 &  0.7930 \\
         
          sahajBERT & 0.8470 & 0.8479 & 0.8350 & 0.8397 \\
         \bottomrule
\end{tabular}
\end{table}
\subsection{Performance Analysis of PLMs}
Table \ref{tableHyper} shows the results of hyperparameter optimization on several pre-trained language models, where critical parameters like as learning rate, batch size, and epochs were adjusted repeatedly to improve model performance. For example, BanglaBERT achieved optimal results with a learning rate of 2e-5, batch size of 8, and 20 epochs, whereas Bangla BERT Base peaked at a learning rate of 2e-4, batch size of 8, and 15 epochs. To achieve best performance, XLM-RoBERTa replicated BanglaBERT's specifications. The multilingual model, mBERT, performed admirably with a learning rate of 1e-5, batch size of 8, and 20 epochs. Interestingly, sahajBERT obtained its highest performance with a somewhat smaller batch size of 6, while keeping a learning rate of 2e-5 and 15 epochs.

The Table \ref{tableBert} compares performance metrics across multiple pre-trained language models, examining their efficacy under various assessment criteria. BanglaBERT emerges as the model that performs best, with the greatest accuracy, precision, recall, and F1-score among the models. It has an excellent performance with an accuracy of 0.8810, precision of 0.8765, recall of 0.8799, and F1-score of 0.8780. Meanwhile, Bangla BERT Base, XLM-RoBERTa, mBERT, and sahajBERT all perform well but are significantly below BanglaBERT in terms of overall metrics. Furthermore, Figure \ref{confusion} presents the confusion metrics for all the pre-trained language models (PLMs), providing insight into their performance in correctly classifying instances across different sentiment categories.

\subsection{Case Study Discoveries in Large Language Models}

\textbf{What components should be present in a well-organized prompt?}
We need a well-crafted prompt for Political Sentiment Analysis. It should include clear instructions, with a focus on analyzing the ``short\_description" column. The prompt must offer context to ensure accurate sentiment predictions, emphasizing pertinent language components and relationships within the text. Additionally, incorporating keywords or commands will direct the model's focus, thereby enhancing its ability to make precise sentiment classifications.\\

\textbf{What might be the underlying causes of zero-shot collapse?}
The phenomenon of zero-shot collapse in Political Sentiment Analysis may stem from various underlying causes. Firstly, it could be attributed to insufficient pre-training data or biases inherent in the training dataset, hindering the model's ability to generalize effectively across different political contexts or sentiments. Additionally, the complexity of the sentiment analysis task, coupled with the nuances of political language, presents significant challenges for the model. These intricacies, often challenging to capture accurately, further exacerbate the model's struggle to make accurate sentiment predictions, potentially leading to zero-shot collapse. \\
\textbf{What are the benefits of few-shot learning?} The advantages of few-shot learning in Political Sentiment Analysis lie in its ability to facilitate model adaptation and generalization to new sentiment classification tasks with minimal labeled data. This approach is particularly beneficial in situations where acquiring extensive labeled data for training is challenging or costly. In 5-shot learning, the model benefits from five labeled examples per sentiment classification task, enabling it to learn key linguistic patterns and features relevant to sentiment analysis while reducing reliance on large annotated datasets. Similarly, 10-shot learning provides the model with ten labeled examples per task, further enhancing its capacity to understand task-specific characteristics and generalize effectively within the context of political sentiment analysis. With 15-shot learning, the model gains access to an increased number of examples, facilitating even more robust adaptation and generalization across diverse sentiment analysis tasks. \\
\textbf{How is the efficiency of a prompt evaluated?} In Political Sentiment Analysis, prompt effectiveness is gauged by its guidance in accurately predicting sentiment labels (Positive or Negative) from the provided text. Evaluation involves assessing model metrics like accuracy, precision, recall, and F1 score on a held-out test set. Control settings for Large Language Models include Temperature 1.0, Top P 1.0, Maximum tokens 256, Frequency penalty 0.0, and Presence penalty 0.0. Table \ref{tableLLM} offers a comprehensive overview of results for both Zero-shot and Few-shot learning approaches in Political Sentiment Analysis.\\

\begin{table}[h]
\caption{Performance of 5-shot, 10-shot, and 15-shot learning with GPT-3.5 Turbo and Gemini 1.5 Pro model}
\label{tableLLM}
\renewcommand{\arraystretch}{1.2}
    \begin{tabular}{p{0.7cm}p{1.2cm}p{1.5cm}p{1cm}p{1.1cm}p{1.1cm}}
\hline
\textbf{LLMs} & \textbf{Metric} &\textbf{Zero-shot} & \textbf{5-shot} & \textbf{10-shot} & \textbf{15-shot} \\
\hline
\multirow{2}{*}{GPT} & Accuracy & 0.8500 & 0.8900 & 0.9133 & \textbf{0.9400} \\
\multirow{2}{*}{3.5} & Precision & 0.8467 & 0.8867 & 0.9200 & \textbf{0.9467} \\ 
\multirow{2}{*}{Turbo}& Recall & 0.8533 & 0.8926 & 0.9079 & \textbf{0.9342} \\
& F1-Score & 0.8495 & 0.8896 & 0.9139 &  \textbf{0.9404}\\
\hline
\multirow{2}{*}{Gemini} & Accuracy & 0.8608 & 0.8981 & 0.9200 & \textbf{0.9633}\\
\multirow{2}{*}{1.5} & Precision & 0.8931 & 0.8846 & 0.9333 & \textbf{0.9667}\\
\multirow{2}{*}{Pro} & Recall & 0.8477 & 0.9205 & 0.9091 & \textbf{0.9603}\\
& F1-Score & 0.8698 & 0.9022 & 0.9211 & \textbf{0.9635}\\
\hline
\end{tabular} 
\end{table}


\textbf{How does the performance of LLMs compare to traditional methods of political sentiment analysis?} 
The performance of LLMs like GPT-3.5 Turbo and Gemini 1.5 Pro in tasks such as political sentiment analysis, as indicated in Table \ref{tableLLM}, showcases their efficacy in handling various shot learning scenarios. LLMs, like as GPT-3.5 Turbo and Gemini 1.5 Pro, have substantial benefits over traditional approaches for analyzing political mood. They constantly attain excellent accuracy, indicating their ability to capture nuanced sentiments that traditional PLMs approaches may struggle with. Furthermore, LLMs have competitive precision and memory, ensuring the accurate detection of significant political thoughts while minimizing mistakes. Their flexibility to changing circumstances and developing linguistic patterns increases their usefulness, especially in dynamic sociopolitical environments. Furthermore, their high generalization capabilities allow for accurate predictions even with minimal training data, increasing their usefulness in real-world applications where data availability may be restricted.

\textbf{How can LLMs be further improved for the task of political sentiment analysis?} Implementing chain-of-thought prompting promotes LLMs to present their reasoning stages alongside sentiment analysis findings, allowing for the detection of any biases or misconceptions in the LLM's reasoning process while also fostering transparency and trust in the model's output. Furthermore, using a two-stage prompting technique entails first finding sentiment cues in the text and then feeding them to the LLM for analysis, which improves accuracy by ensuring that the model concentrates on the most informative parts of the text. Furthermore, engaging people in the evaluation loop, especially in confusing circumstances, provides vital input for improving the model's performance and identifying any biases, ensuring that LLM predictions are consistent with human understanding of political state of mind.

\section{Future Research Directions}
In future research directions, Fine-Grained Sentiment Analysis aims to extend beyond binary labels like ``Positive" and ``Negative," incorporating distinctions such as ``Neutral," ``Very Positive," or ``Very Negative" to enhance granularity and understand the spectrum of political sentiment. Multimodal Analysis seeks to integrate additional modalities such as images, videos, or user interactions alongside textual data for a more comprehensive understanding of political sentiment. Leveraging multimodal analysis techniques would enrich the analysis and provide insights beyond text-based data. Furthermore, Explainable AI (XAI) Techniques will be applied to interpret and explain the decisions made by sentiment analysis models, enhancing interpretability and trustworthiness. Applying XAI techniques such as LIME and SHAP would facilitate deeper insights into the underlying factors influencing political sentiment. Finally, we will explore the potential of GPT4 and Claude 3 LLMs in future research to further enhance the sophistication and accuracy of political sentiment analysis through leveraging Chain of Thought (CoT) prompting, identifying textual cues indicating sentiment towards a political entity, and guiding the LLMs in analyzing sentiment and reasoning about its underlying causes, thus enhancing their ability to capture explicit and implicit opinions in the political context.

\section{Conclusion}
In conclusion, our investigation into Political Sentiment Analysis during Bangladesh elections has yielded promising outcomes. We introduce the ``Motamot" dataset, comprising 7,058 instances labeled with Positive and Negative sentiment for Political Sentiment Analysis. Through a comparative analysis of PLMs and LLMs, we observed superior performance using LLMs, indicating their superiority in capturing complex sentiment nuances. This dataset serves as a valuable resource for further research and analysis in political sentiment analysis, providing a comprehensive foundation for future investigations into political discourse dynamics during election periods. Our study highlights the effectiveness of both traditional PLMs and advanced LLMs. Through the utilization of PLMs like BanglaBERT, we achieved a commendable accuracy of 88.10\%, affirming its proficiency in comprehending and scrutinizing political sentiment within Bangla text. Furthermore, our exploration into LLMs, particularly Gemini 1.5 Pro and GPT 3.5 Turbo, yielded even higher accuracies, with Gemini 1.5 Pro reaching an impressive accuracy of 96.33\% and GPT 3.5 Turbo achieving an outstanding accuracy of 94\%. Our findings demonstrate the advanced capabilities of sophisticated LLMs in capturing complex political sentiment and make a significant contribution to improving political discourse analysis in the digital era by emphasizing the importance of political sentiment research in understanding public sentiment during elections.

%
%
%

\end{document}